\documentclass[conference]{IEEEtran}

\usepackage{cite}

\ifCLASSINFOpdf
\usepackage[pdftex]{graphicx}
\else
\fi
\usepackage{amsmath}
\usepackage{amsfonts} 
\usepackage{amsthm}
\usepackage{amssymb}

\begin{document}

\title{Uncertainty Propagation in Deep Neural Networks Using Extended Kalman Filtering}

\author{
\IEEEauthorblockN{Jessica S. Titensky}
\IEEEauthorblockA{\textit{Massachusetts Institute of Technology}\\
Cambridge, Massachusetts \\
jessst@mit.edu}
\and
\IEEEauthorblockN{Hayden Jananthan}
\IEEEauthorblockA{\textit{Department of Mathematics}\\
\textit{Vanderbilt University}\\
Nashville, Tennessee\\
hayden.r.jananthan@vanderbilt.edu}
\and
\IEEEauthorblockN{Jeremy Kepner}
\IEEEauthorblockA{\textit{Lincoln Laboratory Supercomputing Center} \\
\textit{Massachusetts Institute of Technology}\\
Lexington, Massachusetts \\
kepner@ll.mit.edu}
}
\maketitle

\begin{abstract}
Extended Kalman Filtering (EKF) can be used to propagate and quantify input uncertainty through a Deep Neural Network (DNN) assuming mild hypotheses on the input distribution. This methodology yields results comparable to existing methods of uncertainty propagation for DNNs while lowering the computational overhead considerably. Additionally, EKF allows model error to be naturally incorporated into the output uncertainty.\\
\end{abstract}

\renewcommand\IEEEkeywordsname{Keywords}

%
\IEEEpeerreviewmaketitle

\section{Introduction}
\let\thefootnote\relax\footnotetext{This material is based upon work supported by the Assistant Secretary of Defense for Research and Engineering under Air Force Contract No. FA8721-05-C-0002 and/or FA8702-15-D-0001. Any opinions, findings, conclusions or recommendations expressed in this material are those of the author(s) and do not necessarily reflect the views of the Assistant Secretary of Defense for Research and Engineering.}

A pre-trained Deep Neural Network (DNN) accepts an input vector $\textbf{x}_0$ and outputs a vector $\textbf{x}_L$. Uncertainty from $\textbf{x}_0$ propagates through the DNN resulting in uncertainty in $\textbf{x}_L$, but there remains a question of exactly how the input uncertainty translates into output uncertainty, as well as the role of model error in that resulting uncertainty. This question tends to come up during confidence scoring in areas such as automatic speech recognition where things like background noise can distort the input signal \cite{abdelaziz2015uncertainty}. More precisely, suppose $\textbf x_0$ is the mean of a multivariate normal distribution with covariance matrix $\mathbf \Sigma_0$. As the DNN acts non-linearly on $\textbf x_0$, it is unlikely that the output distribution will be exactly multivariate normal (Gaussian). However, it can be approximated by a Gaussian and modified later if necessary \cite{garthwaite2005statistical}. So, assuming that our output is a multivariate Gaussian with mean $\textbf x_L$, we want to find the output covariance matrix $\mathbf \Sigma_L$  corresponding to the distribution.

Previous approaches for propagating the uncertainty include finding closed form solutions and then numerically integrating probability distributions in the number of hidden nodes dimensions \cite{lee1994input}, which is unrealistic to compute when the DNN has large or many hidden layers. More recently, Monte Carlo sampling and the unscented transform have been used to take a set of samples from the input distribution, propagate them through the DNN, and approximate the first and second moments of the output distribution from them \cite{abdelaziz2015uncertainty}. This can be done to propagate through the DNN as a whole or layer-by-layer, approximating the activation function with a piecewise exponential \cite{astudillo2011propagation}. This method requires sending at least dim$(\textbf x_0)+1$ (where dim gives the dimension of a vector) samples \cite{julier2002reduced} through the DNN for each input we wish to propagate the error for, which is also computationally expensive. Additionally, current methods only find the error in the output which originated directly from the input, not accounting for the inherent error in the DNN itself. In other words, they assume that the DNN is a perfect model, which is rarely the case. Extended Kalman Filtering (EKF) \cite{julier1997new} has already been applied to DNNs, but was done so as a part of the model training process \cite{haykin2004kalman}. Using EKF for uncertainty propagation through DNNs, we can replicate the results yielded by current methods with much less computation, and also account for the model error of the DNN.

\section{Approach}

EKF examines a nonlinear system with a discretized time domain. At each time-step it makes a prediction, using the process noise, control input, and the previous step's state. Using this prediction along with the observation noise, EKF then estimates the system's current state along with the accuracy of that estimation. By treating the layers of the DNN as discrete time steps and their values as states, EKF may be applied. Our system then has no control input and only has observation noise in the first layer in the form of $\mathbf \Sigma_0$. As such, only the prediction step of EKF need be applied.

Take our DNN (Fig. 1)
\begin{figure}
\centering
\includegraphics[scale=0.25]{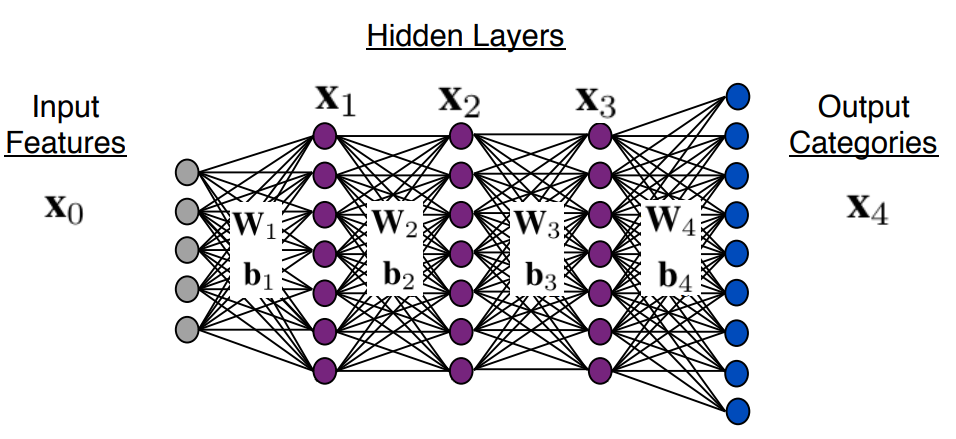}
\caption{Example DNN where $L=4$.}
\end{figure}
to have $L-1$ hidden layers, and say $\textbf x_\ell$ is the vector representing the state estimate at layer $\ell$ and $\mathbf \Sigma_\ell$ is the covariance matrix such that
\begin{align*}
\mathbf \Sigma_\ell(i,j)&=\text{cov}(\textbf x_\ell(i),\textbf x_\ell(j))\\
&=E[\textbf x_\ell(i)\textbf x_\ell(j)]-E[\textbf x_\ell(i)]E[\textbf x_\ell(j)]
\end{align*}
 where $E[X]$ is the expected value of $X$. The nonlinear operation that takes us from state $\ell$ to $\ell+1$ is given by $$\textbf x_{\ell+1}=f(\textbf W_{\ell+1} \textbf x_\ell +\textbf b_{\ell+1})$$ where $f(\textbf x)(i)=\text{max}(0,\textbf x(i))$, $\textbf W_\ell$ is the matrix such that $\textbf W_\ell(i,j)$ is the weight of the edge connecting the $i$-th node in layer $\ell$ to the $j$-th node in layer $\ell-1$, and $\textbf b_\ell$ is the bias vector for layer $\ell$. \\
Note that we specifically assume the use of the Rectified Linear Unit (ReLU) as our activation function $f$. This is well-suited for EKF, which linearizes about the current state's estimate since everywhere (except the point where it's non-differentiable) in the Taylor expansion of ReLU, all the terms after the linear term are 0 anyway. 

The process noise of our system comes from the error resulting from the weights and biases of the pre-trained DNN and for each layer $\ell >0$ is represented by $\textbf Q_\ell$. It can be approximated by the sample covariance matrix which is found by taking a sufficiently large data set of $N$ inputs (separate from the training and testing data sets) and running them through the DNN so then $$\textbf Q_\ell(i,j)=\frac{1}{N-1}\sum_{k=1}^N{(\textbf x_{\ell_k}(i)-\overline{\textbf x_\ell}(i))(\textbf x_{\ell_k}(j)-\overline{\textbf x_\ell}(j))}$$ where $\overline{\textbf x_\ell}(i)=\frac{1}{N}\sum_{k=1}^N{\textbf x_{\ell_k}(i)}$. And equivalently, 
\begin{align*}
\textbf Q_\ell=\frac{1}{N-1}([\textbf x_{\ell_1} \ &\textbf x_{\ell_2} \ \cdots \  \textbf x_{\ell_{N}}]-\overline{\textbf x} \mathbf{1}_{N}^\top)\\
&([\textbf x_{\ell_1} \ \textbf x_{\ell_2} \ \cdots \  \textbf x_{\ell_{N}}]-\overline{\textbf x} \mathbf{1}_{N}^\top)^\top
\end{align*} 

Let $\textbf F_\ell=\nabla_{\textbf x_{\ell-1}}\textbf x_\ell$ be the Jacobian matrix such that 
\begin{align*}
\textbf F_\ell(i,j)&=\frac{\partial \textbf x_\ell(i)}{\partial \textbf x_{\ell-1}(j)}\\
&=\frac{\partial f(\sum_k{\textbf W_\ell(i,k) \textbf x_{\ell-1}(k)} +\textbf b_\ell(i))}{\partial \textbf x_{\ell-1}(j)}\\
&=\frac{\partial f(\sum_k{\textbf W_\ell(i,k) \textbf x_{\ell-1}(k)} +\textbf b_\ell(i))}{\partial \sum_k{\textbf W_\ell(i,k) \textbf x_{\ell-1}(k)} +\textbf b_\ell(i)} \\
&\qquad \cdot \frac{\partial \sum_k{\textbf W_\ell(i,k) \textbf x_{\ell-1}(k)} +\textbf b_\ell(i)}{\partial \textbf x_{\ell-1}(j)}\\
& = \begin{cases} \textbf{W}_\ell(i,j) & \text{if $\textbf{x}_\ell(i) > 0$} \\ 0 & \text{otherwise} \end{cases}
\end{align*}

In most applications of EKF, finding Jacobians dominates the computation time \cite{julier1995new}. Here, however, this is not the case since the $\textbf F_\ell$'s can be computed layer-by-layer from the weight matrices alongside the $\textbf x_\ell$'s.

Finally, we can use the prediction step EKF equations to find the state estimates and covariances for each layer $\ell>0$. These are simply 
\begin{align*}
\textbf x_{\ell}&=f(\textbf W_{\ell} \textbf x_{\ell-1} +\textbf b_{\ell})\\
\mathbf \Sigma_\ell&=\textbf F_\ell \mathbf \Sigma_{\ell-1} \textbf F_\ell^\top +\textbf Q_\ell
\end{align*}
Iteratively applying these until layer $L$ results in the output vector $\textbf x_L$ and its covariance matrix $\mathbf \Sigma_L$. $\mathbf \Sigma_L$ can then be used to find the hyperellipsoid centered at $\textbf x_L$ for a certain confidence level. Alternatively, assuming the components of $\textbf x_L$ are relatively uncorrelated, just $\mathbf \Sigma_L$'s main diagonal can be used to find error bars of a certain confidence level for each component of the output vector independently.

\section{Experimental Results}

We use the MNIST handwritten digit data \cite{lecun1998mnist}, where 28x28 pixel input images (Fig. 2) are converted into 784-dimensional input vectors where each component is between 0 and 1 and the output vectors are 10-dimensional in which each component is nonnegative and represents how likely it is that that digit was the one written. A DNN with 5 hidden layers of 256 nodes each was trained on 50000 images to 92.8\% accuracy and another 10000 images were used to compute the covariance matrices. $\textbf x_0$ (Fig. 2), another image vector distinct from the training and testing image vectors, whose digit label is 9,
\begin{figure}
\centering
\includegraphics[scale=0.5]{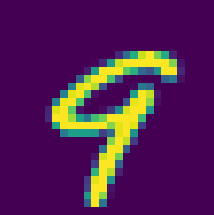}
\caption{Input image.}
\end{figure}
is assigned the diagonal covariance matrix $\mathbf \Sigma_0=.0025\textbf I_{784}$ (so that the components are independent and each has a standard deviation of .05).

Using EKF, $\textbf x_L$ and $\mathbf \Sigma_L$ (Fig. 3) are found
\begin{figure}
\centering
\includegraphics[scale=0.5]{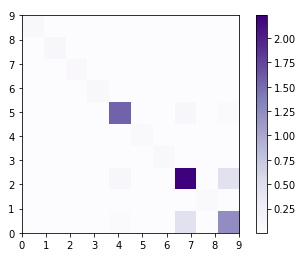}
\caption{Output covariance matrix $\mathbf \Sigma_L$.}
\end{figure}
and because the dominant terms of $\mathbf \Sigma_L$ are along the diagonal, the components of the predicted $\textbf x_L$ can be approximated to be uncorrelated with variances $\sigma^2$'s given by the entries on the main diagonal. Then $1\sigma$ error bars can be plotted against the predicted values to show the confidence region accounting for the original input uncertainty as well as the error provided by the model itself (Fig. 4).
\begin{figure}
\centering
\includegraphics[scale=0.5]{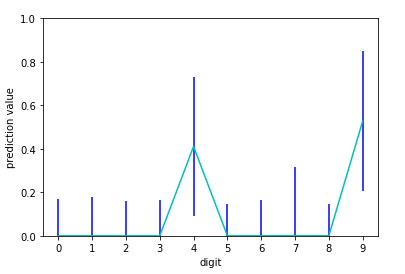}
\caption{$\textbf x_L$ prediction value of each digit (light blue) along with $1\sigma$ error bar (dark blue) on each prediction.}
\end{figure}
For this specific data set, since the components of the output vector must all be nonnegative, each variance was scaled to be that of a truncated normal distribution on [0,$+\infty$) instead of an unbounded normal distribution and correspondingly, the error bars below 0 were cut off.

Repeating this procedure but without adding the $\textbf Q_\ell$ at each layer, so $$\mathbf \Sigma_\ell=\textbf F_\ell \mathbf \Sigma_{\ell-1} \textbf F_\ell^\top$$ we get error bars that depend only on the input uncertainty, effectively assuming that the model is perfect. We can test this by taking a sample of 5000 input image vectors where the components are drawn from independent normal distributions of variance .0025 and centered at the components of $\textbf x_0$, finding the model prediction for each of these samples and then comparing our computed standard deviations for each value of the prediction with the sample standard deviations. We find that the EKF method gives a very similar result to that of the Monte Carlo simulation (Fig. 5).
\begin{figure}
\centering
\includegraphics[scale=0.5]{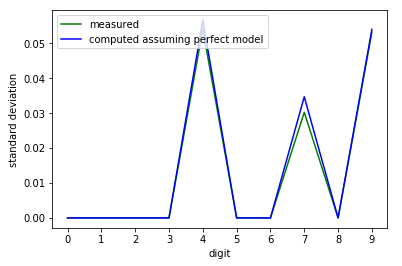}
\caption{Standard deviations computed by EKF assuming a perfect model and sample standard deviations found by Monte Carlo simulation.}
\end{figure}

Without assuming a perfect model, it is difficult to test the accuracy of the resulting error bars due to the inherent error of the model, so the actual standard deviations resulting from the single sample cannot be verified. However, the accuracy can be estimated using an aggregate Root Mean Squared Error (RMSE) calculated by inference testing labeled images with the same label as $\textbf{x}_0$. This RMSE can be compared with the estimated standard deviation calculated by EKF with $\mathbf \Sigma_0= \mathbf{0}$ (Fig. 6). As the EKF-estimated standard deviations represent the accumulation of error through all of the layers while the RMSE only indicates the average error in the final layer, the RMSE will generally be less than the EKF-estimated standard deviations. The effect of a single hidden layer on the error cannot be directly tested because there is no way of knowing what the output of a hidden layer should be. Additionally, that no single image will correctly serve as the 'typical' image for a given label makes the RMSE an even rougher approximation of the real standard deviation. 
\begin{figure}
\centering
\includegraphics[scale=0.5]{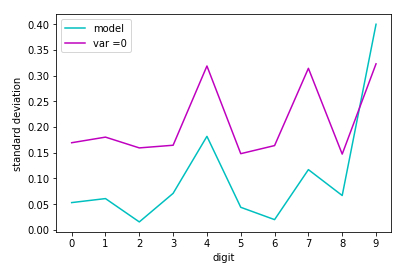}
\caption{Standard deviations computed by EKF with no input error without assuming a perfect model, and root mean squared errors found from the model.}
\end{figure}

Varying the diagonal entries of $\mathbf \Sigma_0$ and comparing the EKF output to the actual standard deviations (assuming a perfect model) or RMSE (without assuming a perfect model) illustrates the relationship between $\mathbf \Sigma_0$ and $\mathbf \Sigma_L$ under those disjoint hypotheses (Fig. 7). Note that here, the higher variances are used for illustrative purposes only and are not likely to reflect actual use-cases as the DNN was trained to expect the components of $\textbf x_0$ to strictly be in the range [0,1].
\begin{figure}
\centering
\includegraphics[scale=0.5]{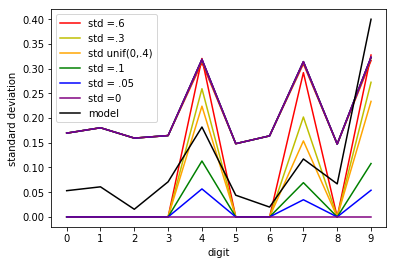}
\caption{Standard deviations computed by EKF starting with the indicated variances along the main diagonal of $\mathbf \Sigma_0$, along with the RMSE found from the model. The top curve is an overlap of all 6 when not assuming a perfect model. The other 6 assume a perfect model.}
\end{figure}

\section{Discussion}

Fig. 7 indicates that when assuming a perfect model, higher input error gives higher output error where ReLU doesn't vanish, and 0 where it does. Additionally, when the input vector component distributions are independent (as assumed in our calculations), the output error plot has the same shape but scales according to the average of the input error. When not assuming a perfect model, the input error plays a very small role in the output error. While the 6 overlapping curves in Fig. 7 are not exactly identical, they only differ from each other by around $10^{-5}$ \textemdash $10^{-2}$. This is because in our model, the $\textbf F$ and $\textbf Q$ matrix entries were very roughly on the orders of around $10^{-2}$ \textemdash $10^{-1}$ and $10^{-5}$ \textemdash $10^{-3}$ respectively, so iteratively scaling by $\textbf F$'s and adding $\textbf Q$'s made the $\mathbf \Sigma$'s tend toward the same values.

Additionally, running this experiment on DNNs with the same topology but trained to different accuracies, we found that the results could be  drastically influenced when using a poorly trained model (Fig. 8).
\begin{figure}
\centering
\includegraphics[scale=0.5]{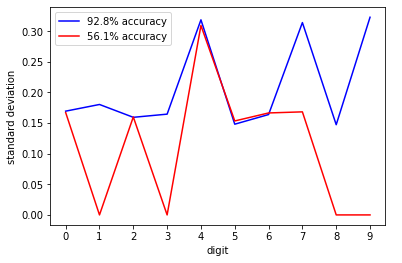}
\caption{Standard deviations computed by EKF without assuming a perfect model with the same $\textbf x_0$ and $\mathbf \Sigma_0 =.0025\textbf I$ on DNNs with identical topologies but trained to different accuracies.}
\end{figure}
In the model trained to 56.1\% accuracy, whose only nonzero prediction value was on digit 4 (as well as other models trained to relatively low accuracies), the variances for some digits are always 0 regardless of input. This is because if the weights or biases are too small, the values of some nodes vanish identically after applying ReLU, zeroing out the $\textbf F$ and $\textbf Q$ terms there as well.

\section{Conclusion}

When assuming a perfect model, using EKF for uncertainty propagation through a DNN gives results comparable to that of previous methods, but requires fewer and simpler computations which can be performed alongside inference tests. Additionally, EKF provides information in the case of an imperfect model, combining both the input uncertainty and the error from the DNN itself to give a more accurate representation of the total uncertainty of the output.  Future work in this area will explore applying EKF to sparse deep neural networks.  The methodology of sparsification includes Hessian-based pruning \cite{lecun1990optimal,hassibi1993second}, Hebbian pruning  \cite{srivastava2014dropout}, matrix decomposition \cite{7298681}, and graph techniques \cite{KepnerGilbert2011,kepner2017enabling,kepner_exact,kumar2018ibm,kepner2018mathematics}, which should be amenable to the EKF approach.

\nocite{julier1996general}
\nocite{kepner2018sparse}
\nocite{julier1995new}
\nocite{amemiya1973regression}
\nocite{kalman1960new}
\nocite{kepner2018mathematics}
\nocite{hinton2012deep}



\section*{Acknowledgments}
%
%

The authors wish to acknowledge the following individuals for their contributions and support:
William Arcand, Bill Bergeron, David Bestor, Bob Bond, Chansup Byun, Alan Edelman, Vijay Gadepally, Chris Hill, Michael Houle, Matthew Hubbell, Michael Jones, Anna Klein, Charles Leiserson, Dave Martinez, Peter Michaleas, Lauren Milechin, Paul Monticciolo, Julia Mullen, Andrew Prout, Antonio Rosa, Albert Reuther, Siddharth Samsi, and Charles Yee.




\bibliographystyle{IEEEtran}
\bibliography{refs} 

%

\end{document}